\documentclass{ifacconf}

\usepackage{graphicx}      
\usepackage{natbib}        
\usepackage{amsmath,amssymb,amsfonts}
\usepackage{algorithmic}
\usepackage{graphicx}
\usepackage{textcomp}
\usepackage{xcolor}
\usepackage{comment}
\usepackage{bm}
\usepackage{booktabs}
\usepackage{makecell}
\usepackage{url}
\usepackage{mathtools}
\usepackage{eso-pic}
\usepackage{xcolor}

\AddToShipoutPictureBG*{%
  \AtPageUpperLeft{%
    \hspace{0.5\paperwidth}%
    \raisebox{-1.5cm}{%
      \makebox[0pt][c]{\textbf{\color{gray} Preprint. Submitted to IFAC World Congress 2026.}}%
    }%
  }%
}
\begin{document}
\begin{frontmatter}

\title{NMPC-based Motion Planning with Adaptive Weighting for Dynamic Object Interception\thanksref{footnoteinfo}} 
\thanks[footnoteinfo]{This work has been submitted to the IFAC World Congress for possible publication. Under review.}

\author[RPTU]{Chen Cai} 
\author[RPTU]{Saksham Kohli} 
\author[RPTU]{Steven Liu}

\address[RPTU]{Department of Electrical and Computer Engineering, University of Kaiserslautern-Landau, 67663 Kaiserslautern, Germany \\
 e-mails: \{chen.cai, kohli, steven.liu\}@rptu.de}

\begin{abstract}                
   This paper presents a nonlinear Model Predictive Control (MPC) planner for dynamic object interception using cooperative manipulator systems under closed-chain constraints. We introduce an Adaptive-Terminal (AT) formulation that employs cost shaping to mitigate actuator power violations common in Primitive-Terminal (PT) approaches. Experimental validation on a physical platform demonstrates superior motion quality and robustness compared to the PT baseline. Crucially, the system exhibits excellent real-time performance, achieving an average computation time of $19\,\text{ms}$--less than half the $40\,\text{ms}$ sampling interval. This establishes the framework's suitability for agile, safety-critical cooperative tasks.
\end{abstract}

\begin{keyword}
   Motion Planning, Multi-Robot Systems, Model Predictive Control (MPC), Robotic Catching
\end{keyword}

\end{frontmatter}
\section{Introduction}
Catching fast-moving objects mid-air is a canonical benchmark for robotic agility and coordination. This demanding task necessitates closing the perception-prediction-planning-control loop in real-time under tight timing constraints. Beyond its fundamental research value, dynamic interception finds new relevance in collaborative manufacturing, embodying Industry 5.0 principles. Such non-contact transfer methods exemplify technology that supports rather than constrains human workers, contrasting with traditional handover interfaces requiring precise spatial alignment or physical contact.

This work employs two coordinated manipulators, abstracting the human upper body to explore dynamic behaviors like catching--signatures of embodied intelligence. Functionally, this cooperative setup enables richer interception geometries, secure cradle-like grasps, enhancing robustness and safety. However, coordination introduces significant challenges: closed-chain geometry, inter-arm clearance, and kinematic redundancy, making control substantially more complex than single-arm cases. Addressing dynamic catching on such platforms is thus not merely harder, but a necessary step toward foundational, human-like intelligent behaviors for next-generation humanoid and collaborative robots.

To achieve dynamic ball catching with coordinated manipulators, we propose an integrated perception-prediction-planning-control pipeline, experimentally validated. At its core lies an NMPC-based motion planner generating coordinated trajectories for the closed-chain system. Our approach explicitly enforces kinematic consistency and joint limits. Operating within a predefined safety envelope circumvents complex analytical workspace characterization. Dynamic interception demands balancing rapid convergence with motion smoothness for stability and constraint adherence. Addressing this trade-off, we introduce an Adaptive-Terminal (AT) formulation. AT dynamically modulates pose tracking cost weights based on state error. This implicitly regularizes control effort for smoother joint velocities, critically mitigating power limit violations-a frequent challenge with aggressive, fixed-weight approaches in such systems. Driven by a single depth camera and a Kalman filter predictor, system efficacy is validated via simulation and experiments, quantifying success rates, latency, and coordination fidelity (e.g., error in maintaining the fixed grasping distance).

\section{Related Work}
Research in the domain of robotic catching in flight has established a common methodological pipeline \citep{frese2001off}, typically involving four sequential stages: 1) predicting the object's trajectory based on sensor measurements, 2) selecting a feasible interception point and computing the corresponding desired robot posture 3) planning a time-critical motion to reach this posture for a successful catch, and 4) motion control to execute the planned trajectories. While this framework has been extensively explored for single-arm manipulators \citep{bauml2010kinematically}, the research landscape is evolving to address more complex platforms such as mobile manipulators \citep{dong2020catch} and humanoids \citep{kober2012playing, bauml2011catching}, which necessitate whole-body coordination.

Methodologically, current approaches are broadly categorized into model-based optimal control and data-driven learning paradigms. Learning-based methods, ranging from imitation learning using human demonstrations to reinforcement learning (RL) \citep{abeyruwan2023sim2real}, can learn complex dynamics and adapt to real-world physical nuances \citep{yu2021neural}. For instance, \citet{kim2014catching} utilized a learning-by-demonstration approach, combining probabilistic models for grasp configurations with dynamical systems (DS) to generate coordinated arm-hand motions for catching irregularly shaped objects. However, these methods are often requiring extensive data and may struggle with robustness when faced with significant distributional shifts beyond their training data \citep{marwan2021comprehensive}. On the other hand, model-based approaches, particularly Model Predictive Control (MPC), offer strong real-time performance, zero-shot adaptation to new targets, and explicitly constraints handling, although they are sensitive to modeling errors. A recent comprehensive study by \citet{abeyruwan2023agile} compared a whole-body MPC controller with a blackbox learning policy, concluding that while the fine-tuned learning agent achieved higher peak performance, the MPC-based agent demonstrated substantially greater robustness to out-of-distribution throws.

While the aforementioned methods advance robotic agility, they predominantly focus on single-arm systems. Furthermore, to achieve a successful catch, these works often replace the robot's standard gripper with a specialized, custom-designed end-effector, such as a Lacrosse head or a bespoke catching tool. Our work, in contrast, is motivated by the objective of using standard, commercial-off-the-shelf (COTS) manipulators with their original, general-purpose grippers. As these standard grippers are ill-suited for directly intercepting a ball, a practical alternative is a tool-based strategy: the manipulators grasp a separate, passive catcher (e.g., a container or net). However, stably manipulating such a tool at high accelerations is physically problematic for a single arm. Consequently, a cooperative bimanual grasp is necessary to ensure a stable, multi-point hold on the catcher. While this bimanual, tool-based approach enables robust, "cradle-like" interception using only COTS hardware, it fundamentally transforms the control problem by introducing a rigid closed-chain constraint.

Our work addresses the challenging, yet less explored, problem of dynamic interception using two cooperative manipulators forming a closed kinematic chain. This configuration introduces fundamental difficulties rendering traditional pipeline suboptimal. Firstly, characterizing the cooperative task space and selecting an optimal catching pose a priori becomes largely intractable. The reachable workspace for the jointly held object is a coupled, constrained manifold \citep{he2022distributed}, not merely a union of individual workspaces. Furthermore, combined redundancy means numerous joint configurations can realize a target pose, each with vastly different dynamic implications, making predefined optimal pose selection ill-posed. Secondly, the closed-chain constraint imposes stringent coordination requirements for motion planning and execution, alongside practical deployment challenges. Aggressive interception maneuvers demand significant instantaneous power, frequently triggering power limit violation faults on collaborative platforms. This issue is particularly exacerbated in closed-chain systems, where slight misalignments or tracking errors induce internal forces, increasing power consumption. Explicitly modeling such complex dynamics, like power limits, as constraints within real-time NMPC frameworks remains computationally prohibitive.

To overcome these limitations, we propose a nonlinear MPC framework that holistically addresses the planning problem for cooperative, closed-chain interception. By operating within a safety envelope, our approach circumvents deriving complex analytical models of the cooperative workspace. Based on the predicted ball trajectory, a target selection algorithm identifies a desired interception pose prioritized for kinematic feasibility. 
This pose then serves as the target for the catcher within the NMPC, which subsequently optimizes the coordinated joint trajectories for both manipulators to maneuver the catcher to that pose. This implicitly resolves system redundancy within the optimization horizon based on task objectives and constraints, avoiding the difficult a priori selection of a single optimal system configuration. Furthermore, the introduced Adaptive-Terminal (AT) formulation dynamically adjusts control objectives, providing a practical means to balance rapid interception with the smooth, coordinated motion required to mitigate execution challenges like power limit violations inherent in such demanding dynamic closed-chain tasks.

\section{Problem Formulation}
This section delineates the problem of cooperative robotic ball catching. The system architecture and relevant preliminaries are established first, leading to a formal statement of the motion planning task.

\subsection{System Overview}
\begin{figure}[t]
    \centering
    \includegraphics[width=0.95\columnwidth]{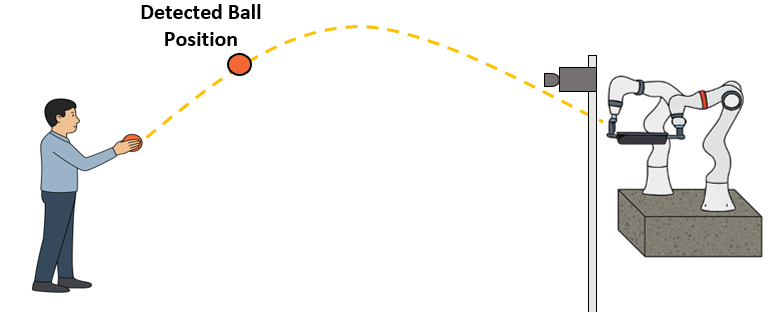}
    \caption{Conceptual overview of the robotic ball-catching task. A perception system observes the thrown ball's trajectory, enabling the cooperative robotic system to plan and execute a coordinated in-flight interception.}
    \label{fig:system_overview}
\end{figure}

The focus of this work is the dynamic task of catching a thrown ball using a system of multiple coordinated robots, as conceptually illustrated in Fig.~\ref{fig:system_overview}. The task involves a human agent or a mechanical thrower launching an object along a ballistic trajectory. A key challenge is the inherent uncertainty of this task: the trajectory is non-deterministic, featuring randomized initial velocities and launch angles that result in interception points within a predefined operational workspace. This variability necessitates a motion planner capable of real-time adaptation and continuous re-planning, rather than executing a static, pre-computed path. The robotic system, composed of two 7-DoF robots rigidly holding a catcher--which in this work takes the form of a container--must visually perceive the object's flight, plan a suitable interception maneuver, and execute a coordinated motion to catch it before it lands.

To formally describe the system, we denote the joint position for the left and right manipulators as $\bm{q}_l(t), \bm{q}_r(t) \in \mathbb{R}^7$, respectively, stacked into the overall system configuration $\bm{q}(t) = [\bm{q}_l(t)^T, \bm{q}_r(t)^T]^T \in \mathbb{R}^{14}$.
Consequently, the system's joint velocities and accelerations are given by $\dot{\bm{q}}(t)$ and $\ddot{\bm{q}}(t)$. The thrown ball's state at time $t$ consists of its Cartesian position $\bm{p}_b(t) \in \mathbb{R}^3$ and velocity $\bm{v}_b(t) \in \mathbb{R}^3$. The predicted velocity at interception, $\bm{v}_b(t_{\text{catch}})$, dictates the primary alignment requirement. This defines the impact direction unit vector $\bm{d}_{\text{impact}} = \bm{v}_b(t_{\text{catch}}) / ||\bm{v}_b(t_{\text{catch}})||$, which, along with task-specific constraints (e.g., keeping the catcher opening upwards in this work), determines the desired catcher orientation represented by the unit quaternion $\phi_b$. 

\subsection{Preliminaries}
The dynamics of each manipulator ($i \in \{l, r\}$) are governed by the standard rigid-body equation of motion \cite{cai2025mpc}:
\begin{equation}
    \bm{M}_i(\bm{q}_i)\ddot{\bm{q}}_i + \bm{C}_i(\bm{q}_i, \dot{\bm{q}}_i)\dot{\bm{q}}_i + \bm{g}_i(\bm{q}_i) = \bm{\tau}_i + \bm{\tau}_{\text{ext}, i}
    \label{eq:robot_dynamics}
\end{equation}
Here,  $\bm{\tau}_i \in \mathbb{R}^7$ is the vector of actuator torques. The matrices $\bm{M}_i$, $\bm{C}_i$, and the vector $\bm{g}_i$ are the configuration-dependent inertia, Coriolis/centrifugal, and gravity terms, respectively.

For real-time tractability, a hierarchical control structure is adopted, separating high-level kinematic motion planning from low-level dynamic control in (\ref{eq:robot_dynamics}). The NMPC motion planner (high-level) operates based on a simplified kinematic model, specifically a double integrator,
\begin{equation}
\ddot{\bm{q}}(t) = \bm{u}(t)
\label{eq: doubleintegrator}
\end{equation}
where the control input $\bm{u}(t) \in \mathbb{R}^{14}$ represents the joint accelerations. This formulation relies on two key assumptions: (i) a low-level controller can accurately track the reference accelerations, and (ii) the dynamics of the lightweight container are negligible. 

\begin{figure}[t]
    \centering
    \includegraphics[width=0.9\columnwidth]{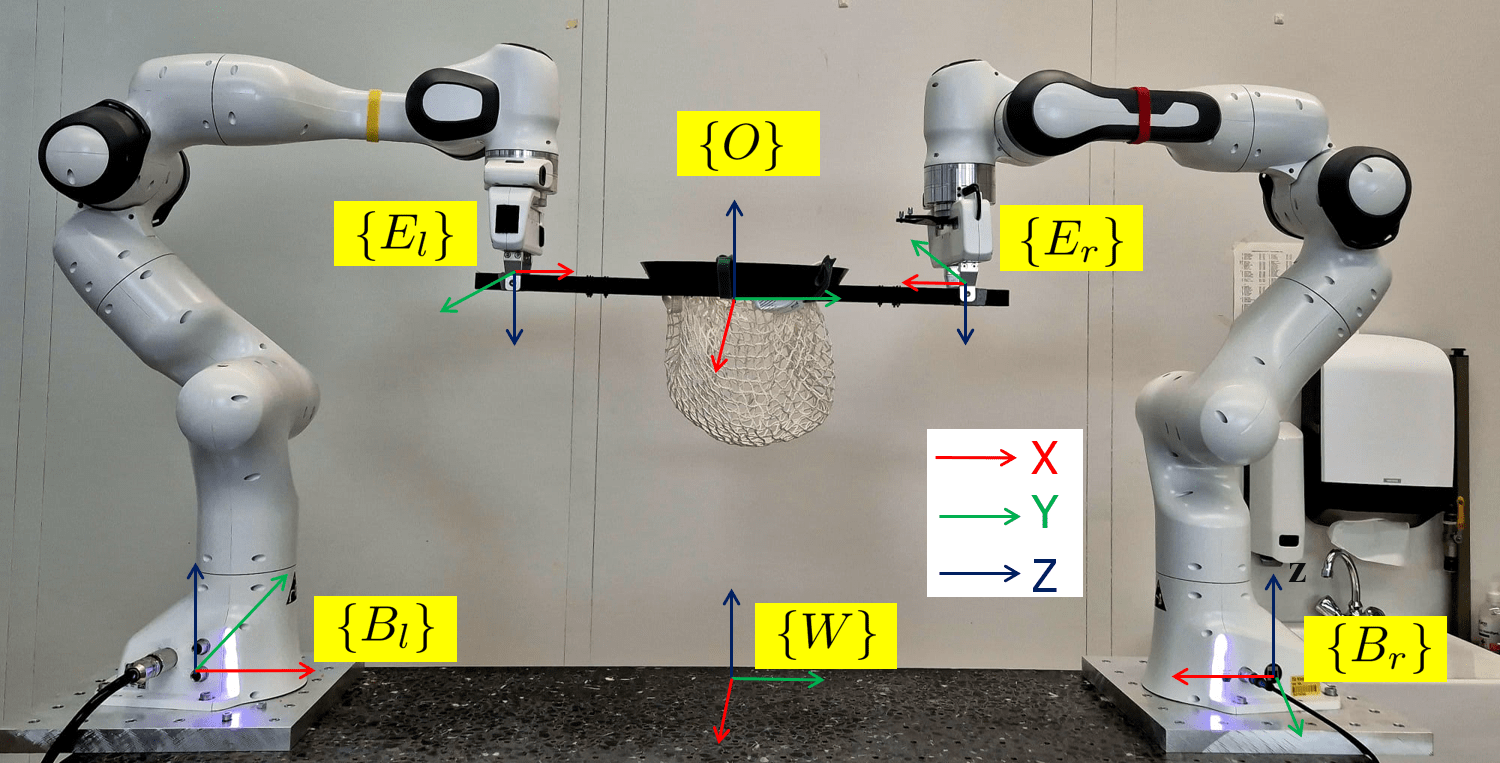}
    \caption{Coordinate system definitions for the multi-robot experimental setup.}
    \label{fig:coord_systems}
\end{figure}

As depicted in Fig.~\ref{fig:coord_systems}, we define a world frame $\{W\}$ and attach frames to the base $\{B_i\}$, end-effector $\{E_i\}$ for each robot $i \in \{l,r\}$, and the catcher $\{O\}$. The forward kinematics mapping, denoted by $\bm{h}_i(\cdot)$, relates the joint configuration $\bm{q}_i$ of an arm ($i \in \{l, r\}$) to the Cartesian pose of its end-effector, $\bm{x}_{E_i} = [\bm{p}_{E_i}^T, \bm{\phi}_{E_i}^T]^T \in SE(3)$, where $\bm{p}_{E_i}$ is the position and $\bm{\phi}_{E_i}$ is the orientation (unit quaternion). The corresponding differential kinematics are given by $\dot{\bm{x}}_{E_i} = \bm{J}_i(\bm{q}_i)\dot{\bm{q}}_i$, where $\dot{\bm{x}}_{E_i}$ is the end-effector twist (linear velocities and quaternion rates) and $\bm{J}_i(\bm{q}_i)$ is the manipulator Jacobian.

The rigid grasp on the container forms a closed kinematic chain. Let the constant transformations from each end-effector frame to the catcher frame be ${}^{E_l}\bm{T}_{O}$ and ${}^{E_r}\bm{T}_{O}$. This allows the catcher pose, $\bm{x}_o = [\bm{p}_o^T, \bm{\phi}_o^T]^T$, to be expressed as a function of either arm's joint configuration through a composite forward kinematics mapping, which we denote as $\bm{h}_o(\cdot)$. This relationship is established via the composition of homogeneous transformations:
\begin{equation}
    {}^{W}\bm{T}_{O} = {}^{W}\bm{T}_{E_i}(\bm{q}_i) \cdot {}^{E_i}\bm{T}_{O}
\end{equation}
The mapping $\bm{x}_o = \bm{h}_o(\bm{q}_i)$ thus represents the forward kinematics from a single arm's joint space to the commonly held catcher task space. As the catcher pose must be unique regardless of which arm is used for computation, this induces the fundamental closed-chain constraint that must hold throughout the motion:
\begin{equation}
    \bm{h}_o(\bm{q}_l) = \bm{h}_o(\bm{q}_r)
    \label{eq:closed_chain_constraint}
\end{equation}
This equality couples the configurations of both arms and defines the feasible manifold where the system operates.

\subsection{Problem Statement}
\label{subsec:ProbState}

The central problem is to compute a motion plan satisfying kinematic constraints and actuator limitations for the cooperative multi-arm system to intercept a thrown ball. Given the initial state $(\bm{q}(t_0), \dot{\bm{q}}(t_0))$ and the ball's predicted trajectory $(\hat{\bm{p}}_b(t), \hat{\bm{v}}_b(t))$, the task is formulated as an Optimal Control Problem (OCP). The objective is to find a control input $\bm{u}(t) = \ddot{\bm{q}}(t)$ that minimizes a cost function $J$:
\begin{equation}
\min_{\bm{u}(t)} J = \Phi(\bm{x}_o(\bm{q}(t_f)), \hat{\bm{p}}_b(t_f), \hat{\bm{\phi}}_b(t_f)) + \int_{t_0}^{t_f} L(\dot{\bm{q}}(t), \bm{u}(t)) dt
\label{eq:ocp_cost}
\end{equation}
where the running cost $L(\cdot)$ penalizes excessive joint velocities and accelerations to promote smooth motions. The terminal cost $\Phi(\cdot)$ should serve as a heavily weighted soft constraint, penalizing the final discrepancy between the catcher pose and the desired interception pose to drive the system towards the goal configuration.

This optimization is subject to the following constraints:
\begin{enumerate}
    \item \textbf{System Dynamics:} The system evolves according to the double integrator model $\ddot{\bm{q}}(t) = \bm{u}(t)$, starting from the given initial state $(\bm{q}(t_0), \dot{\bm{q}}(t_0))$.
    \item \textbf{Actuator Limits:} Joint position, velocity, and acceleration limits must be respected, e.g., $\bm{q}(t) \in [\bm{q}_{\min}, \bm{q}_{\max}]$.
    \item \textbf{Closed-Chain Constraint:} The coordination constraint $\bm{h}_o(\bm{q}_l(t)) = \bm{h}_o(\bm{q}_r(t))$ must be satisfied.
    \item \textbf{Workspace and Self-Collision Constraints:} Motion must remain within a predefined safe workspace. Dedicated constraints ensuring self-collision avoidance (e.g., minimum distance between links) must be satisfied.
\end{enumerate}

Employing a kinematic planner based on Eq.~\eqref{eq: doubleintegrator} offers significant advantages in computational tractability, enabling the real-time performance crucial for dynamic interception tasks. However, the application of this kinematic planning approach within the demanding context of dynamic interception using an inherently stiff, closed-chain system presents significant challenges, particularly as these systems are often rendered dynamically stiff by the high-gain low-level controllers required for precise trajectory tracking. This inherent stiffness means even minor discrepancies between the arms or deviations from the planned path can induce large internal forces. Consequently, there is a heightened risk of triggering actuator power limit violations--critical safety features inherent in most commercial collaborative robots designed to halt motion upon detecting excessive loads or potential collisions. Since incorporating full dynamic constraints to prevent this is often intractable for real-time OCP, a specialized motion planning approach is required to balance rapid task execution with these intrinsic system limitations, as detailed in the following section.

\section{Proposed Motion Planner}
\begin{figure}[t]
    \centering
    \includegraphics[width=0.9\columnwidth]{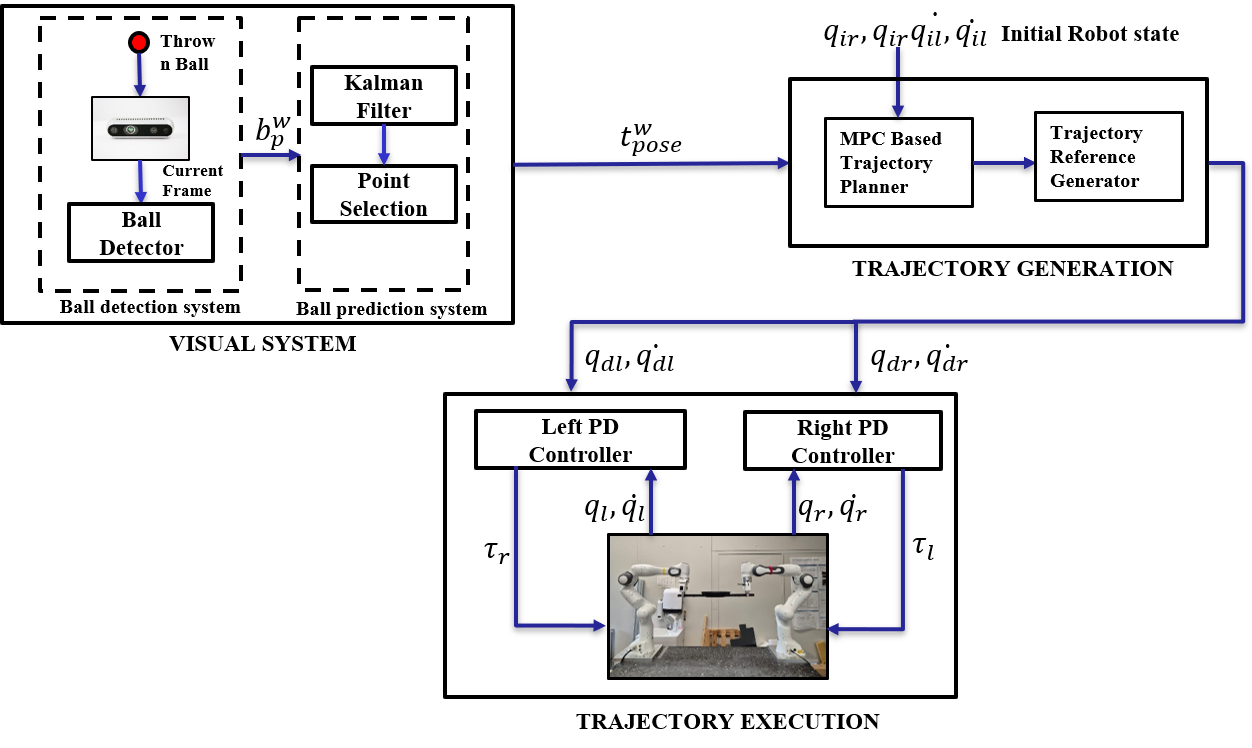}
    \caption{The architecture of the system in the form of diagram}
    \label{fig:system diagram}
\end{figure}

To addressing the closed-chain coordination and dynamic feasibility challenges outlined in Section~\ref{subsec:ProbState}, we propose the motion planning framework depicted in Fig.~\ref{fig:system diagram}. This hierarchical system consists of three main stages: (i) ball trajectory estimation and target point selection, processing visual input to determine a feasible interception pose; (ii) the core nonlinear MPC-based motion planner, generating coordinated joint trajectories towards this target; and (iii) the low-level control layer, responsible for real-time trajectory execution using joint PD controllers. This modular design advantageously decouples planning from execution. Section IV-A details the perception stage, followed by the proposed MPC formulation in Section IV-B.

\subsection{Ball Trajectory Estimation and Target Point Selection}
The perception pipeline initiates by processing RGB-D images via a YOLOv8-based object detector to obtain initial estimates of the ball's 3D position. Robustness against high-speed motion and diverse conditions was ensured through tailored training (details in supplementary material). This stage yields a stream of time-stamped, potentially noisy, 3D position measurements of the ball in the world frame.

These position measurements are subsequently processed by a Kalman filter. 
Employing a discrete-time projectile motion model (including gravity and process noise for unmodeled dynamics), the filter provides smoothed estimates of the ball's full state (position and velocity). Crucially, these current state estimates serve as the initial condition for predicting the ball's future flight trajectory, denoted by $\hat{\bm{x}}_b(t) = [\hat{\bm{p}}_b(t)^T, \hat{\bm{v}}_b(t)^T]^T \in \mathbb{R}^6$, by iteratively propagating the motion model forward. 
This predicted trajectory serves as the input for the subsequent target point selection.

From the predicted trajectory, a target point selection algorithm identifies a feasible interception pose $\bm{p}_{\text{tg}}$ for the subsequent MPC planner. This involves filtering points within a safe workspace envelope $\mathcal{S}$ and selecting the candidate that minimizes the required average velocity for interception, thus balancing responsiveness and kinematic feasibility\footnote{Detailed algorithm and source code available at: \\ \url{https://github.com/Ashray4/nmpc_ada_ballCatching}}
\subsection{Motion Planning}
\subsubsection{Primitive MPC Formulation (PT)}
Based on (\ref{eq: doubleintegrator}), we define the state vector $\bm{z}(k)\! =\! [\bm{q}(k)^T, \dot{\bm{q}}(k)^T]^T \in \mathbb{R}^{2n}$ and the input vector $\bm{u}(k) = \ddot{\bm{q}}(k) \in \mathbb{R}^{n}$, including all $n=14$ joints. Discretized with step $T_s$, system dynamics is described by:
\begin{equation}
    \bm{z}(k+1) = \begin{bmatrix} \bm{I}_n & T_s\bm{I}_n \\ \bm{0}_n & \bm{I}_n \end{bmatrix} \bm{z}(k) + \begin{bmatrix} \frac{T_s^2}{2}\bm{I}_n \\ T_s\bm{I}_n \end{bmatrix} \bm{u}(k)
    \label{eq:mpc_ss_model}
\end{equation}

Based on this model, we first formulate a basic planner, which prioritizes reaching the target interception pose $\bm{x}_{\text{tg}} = [\bm{p}_{\text{tg}}^T, \bm{\phi}_{\text{tg}}]^T$ with high precision. This objective is primarily encoded in a heavily weighted terminal cost. To ensure the motion is smooth and physically feasible, the formulation also includes running costs that penalize excessive joint velocities and accelerations. This leads to the following MPC optimization problem:
\begin{multline}
    \min_{\{\bm{u}(k), \bm{z}(k)\}_{k=0}^{N-1}} J_{\text{PT}} = \| \bm{p}_o(N) - \bm{p}_{\text{tg}} \|^2_{\bm{P}_{e}} +\| \bm{\phi}_{o}(N) - \bm{\phi}_{\text{tg}} \|^2_{\bm{O}_{e}}\\
    +  \sum_{k=0}^{N-1} \left(
     \| \bm{u}(k) \|^2_{\bm{R}} + \| \dot{\bm{q}}(k) \|^2_{\bm{W}}
    \right)
    \label{eq:mpc_cost_primitive}
\end{multline}
subject to:
\begin{align*}
& \text{Eq. } (\ref{eq:mpc_ss_model}) \\ 
& h_{o}(\bm{q}_{l}(k)) = h_{o}(\bm{q}_{r}(k)) = x_o  \\
& \bm{q}_{\min} \leq \bm{q}(k) \leq \bm{q}_{\max}   \\
& \dot{\bm{q}}_{\min} \leq \dot{\bm{q}}(k) \leq \dot{\bm{q}}_{\max}   \\
& \bm{u}_{\min} \leq \bm{u}(k) \leq \bm{u}_{\max}  
\end{align*}

where $\| \bm{\phi}_{o}(N) - \bm{\phi}_{\text{tg}} \|^2_{\bm{O}_{e}}$ represents the weighted squared final orientation error, calculated using a suitable metric for quaternion difference. The large terminal weights $\bm{P}_{e}$ and $\bm{O}_{e}$ drive the optimizer to aggressively minimize the final pose error, while the running cost weights $\bm{R}$ and $\bm{W}$ regulate the control effort profile throughout the trajectory.

\subsubsection{Adaptive-Terminal MPC Formulation (AT)}
The preceding primitive MPC, while effective, utilizes fixed, large terminal weights that can cause aggressive accelerations, velocity overshoots, and potential power limit violations in the stiff closed-chain system. To enhance dynamic feasibility while retaining responsiveness, we propose the Adaptive-Terminal (AT) formulation featuring a cost shaping approach. AT adaptively modulates both the terminal cost weights (cycle-to-cycle) and the stage cost weights (step-by-step within the horizon based on error). This dynamic adjustment allows the planner to balance endpoint precision with trajectory smoothness, leading to the following optimization problem:
\begin{multline}
\!\!\!\!\!\!\!\min_{\{\bm{u}(k), \bm{z}(k)\}_{k=0}^{N-1}}\!\!J_{\text{AT}} \!=\! \| \bm{p}_o(N) - \bm{p}_{\text{tg}} \|^2_{\bm{P}_{e}^{\text{adp}}} +\| \bm{\phi}_{o}(N) - \bm{\phi}_{\text{tg}} \|^2_{\bm{O}_{e}^{\text{adp}}}\\
+  \sum_{k=0}^{N-1} \left(
 \| \bm{p}_o(k) - \bm{p}_{\text{tg}} \|^2_{\bm{Q}_{pos}^{\text{adp}}(k)} + \| \bm{\phi}_{o}(k) - \bm{\phi}_{\text{tg}} \|^2_{\bm{Q}_{ori}^{\text{adp}}(k)}
 \right)
\\
+  \sum_{k=0}^{N-1} \left(
 \| \bm{u}(k) \|^2_{\bm{R}} + \| \dot{\bm{q}}(k) \|^2_{\bm{W}}
 \right)
\label{eq:mpc_cost_adaptive}
\end{multline}
subject to the same constraints as in (\ref{eq:mpc_cost_primitive}).

The core of the AT strategy lies in adaptively shaping the cost function weights based on the error relative to the target pose $\bm{x}_{\text{tg}} = [\bm{p}_{\text{tg}}^T, \bm{\phi}_{\text{tg}}^T]^T$. Both the terminal and stage cost weights employ a shared adaptive law defined by:
\begin{equation}
    W(e; W^{\min}, W^{\max}, \epsilon) = W^{\min} + (W^{\max} - W^{\min}) \cdot \frac{e}{e + \epsilon}
    \label{eq:adaptive_law_general}
\end{equation}
where $e$ is the relevant error magnitude (position $e_{pos} = \| \bm{p}_o(k) - \bm{p}_{\text{target}} \|$ or orientation $e_{ori}$ based on quaternion difference), $[W^{\min}, W^{\max}]$ defines the modulation range for the specific weight, and $\epsilon > 0$ is a sensitivity parameter controlling the transition steepness.

The application of this law differs for stage and terminal weights:
\begin{itemize}
    \item \textbf{Stage Weights} ($\bm{Q}_{pos}^{\text{adp}}(k), \bm{Q}_{ori}^{\text{adp}}(k)$): These are updated at each prediction step $k$ ($k=0, \dots, N-1$) within the horizon. Their values are computed via Eq.~\eqref{eq:adaptive_law_general} within their respective ranges $[\bm{Q}^{\min}, \bm{Q}^{\max}]$ using the predicted error $e(k)$ at that specific future step $k$.
    \item \textbf{Terminal Weights} (${\bm{P}_{e}^{\text{adp}}}, {\bm{O}_{e}^{\text{adp}}}$): These are determined once at the beginning of each MPC cycle, using Eq.~\eqref{eq:adaptive_law_general} based on the initial state error $e(0)$ of the current cycle. The resulting values then remain constant throughout the entire prediction horizon within this cycle.
\end{itemize}
Crucially, the terminal weights operate over a significantly wider modulation range than the stage weights. This design ensures that large initial errors $e(0)$ result in dominant terminal costs, promoting aggressive convergence. Conversely, as the system approaches the target (small $e(0)$ and $e(k)$), the pronounced reduction in terminal weights increases the relative influence of the running costs (regulating $\|\bm{u}\|^2_{\bm{R}}$, $\|\dot{\bm{q}}\|^2_{\bm{W}}$, and stage errors via $\bm{Q}^{\text{adp}}(k)$). This shift naturally favors smoother motion during the final approach, improving feasibility.


\section{Experimental evaluation}
\subsection{Overview}
\begin{figure}[t]
    \centering
    \includegraphics[width=0.9\columnwidth]{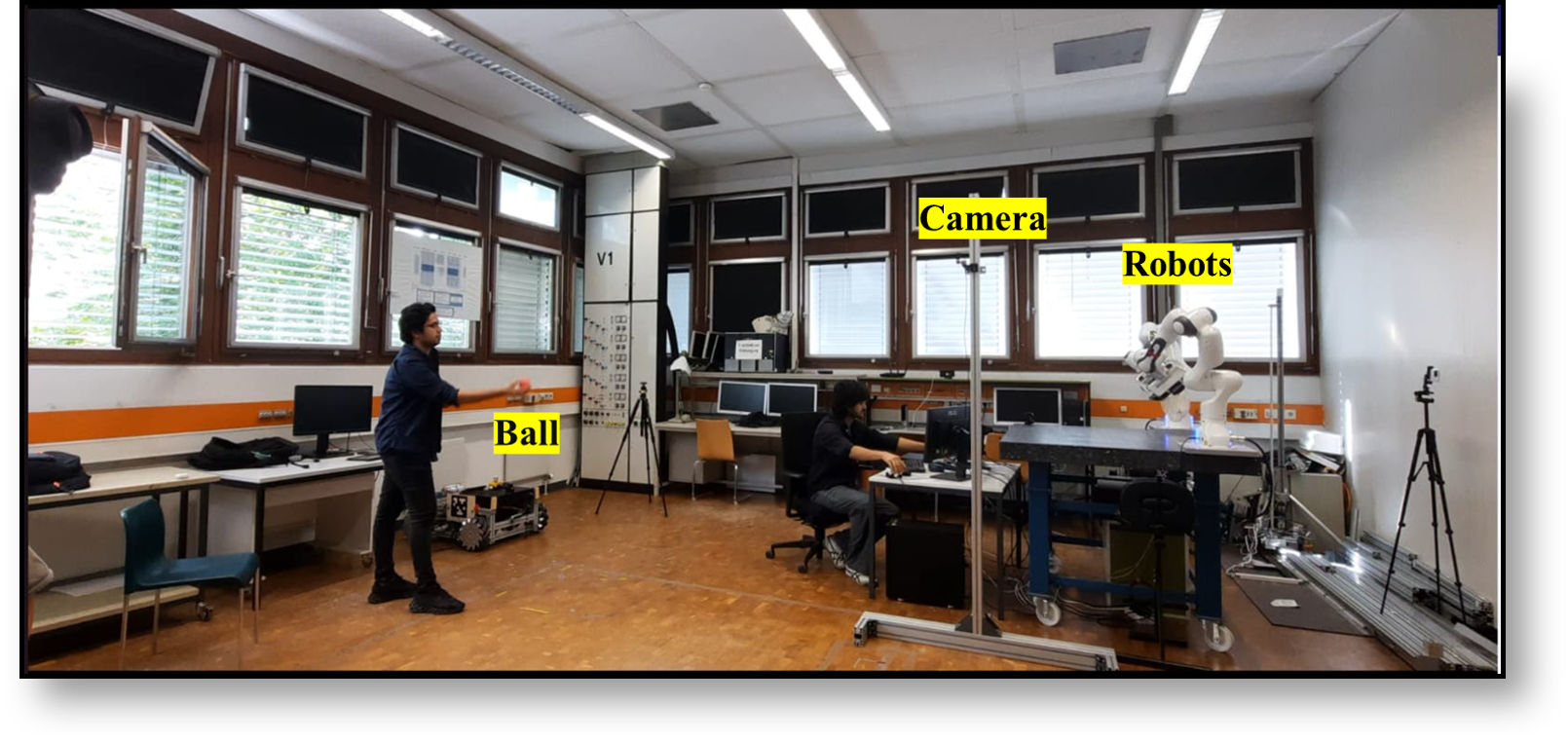}
    \caption{The experimental setup, showcasing the dual Franka Emika Panda manipulators, the cooperative catching container, and the perception system.}
    \label{fig:real_system_setup}
\end{figure}

Experiments were conducted using two 7-DoF Franka manipulators holding a custom 3D-printed container, observed by an extrinsically calibrated Intel RealSense D455 camera (Fig.~\ref{fig:real_system_setup}). A supplementary video is available\footnote{\url{https://youtu.be/LJLC-HzX3G4?si=D9Glj3p1mu6e2AW4}}.

\begin{figure}[t]
    \centering
    \includegraphics[width=0.9\columnwidth]{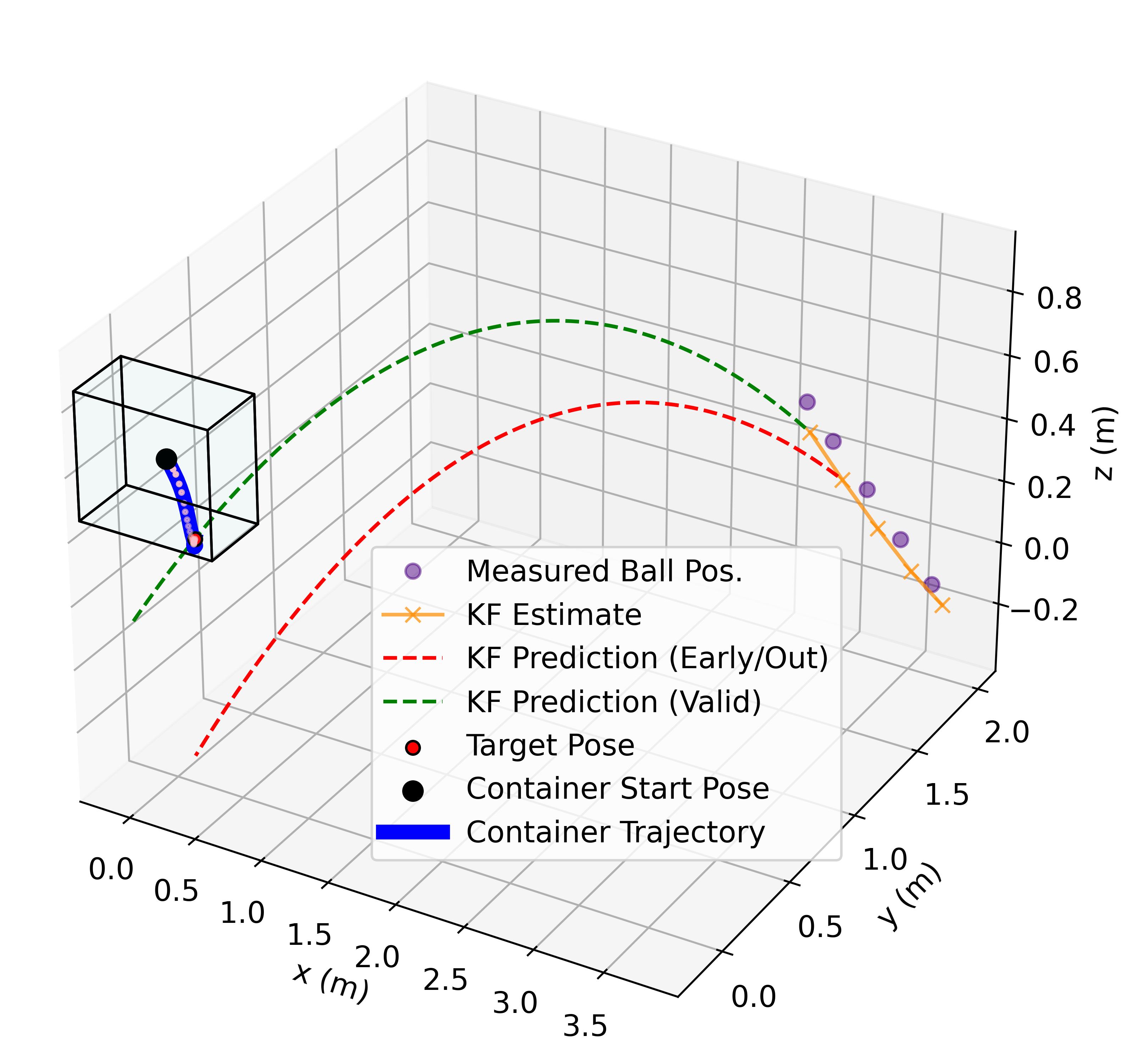}
    \caption{Visualization of a successful ball catch using the AT-MPC framework. 'Measured Ball Pos' (blue markers) are raw vision system detections, filtered into the 'KF Estimate' trajectory (orange line). The Kalman filter generates future path predictions: an early prediction potentially outside the workspace ('KF Prediction (Early/Out)', red line) and a later 'KF Prediction (Valid)' (green dashed line) intersecting the 'Safe Zone' (black cube). This valid prediction determines the 'Target Pose' (red circle) for the container center. The 'Container Trajectory' (bold blue line) shows the executed path of the container center, starting from 'Container Start Pose' (black circle), successfully intercepting the ball.}
    \label{fig:Fig_3D_Catch}
\end{figure}

\begin{figure}[t]
    \centering
    \includegraphics[width=0.9\columnwidth]{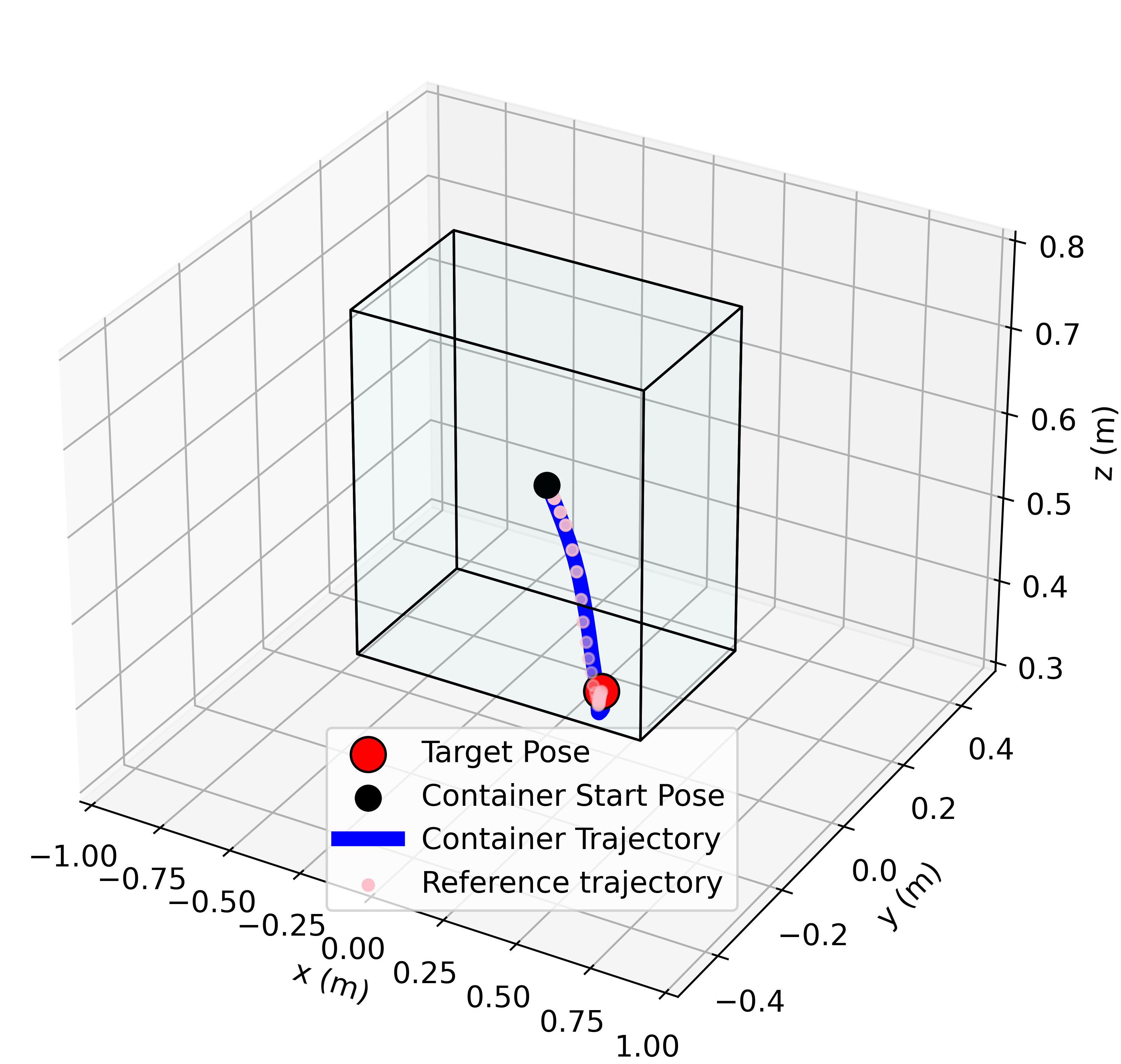}
    \caption{Close Up view}
    \label{fig:Fig_3D_Catch Close}
\end{figure}

MPC-based motion planner and control algorithms executed on a PC (AMD Ryzen Threadripper 3960X, Nvidia RTX A5000) running Ubuntu 22.04 with PREEMPT\_RT patch. The system utilizes ROS 2 Humble and ros2\_control for dual-arm coordination. Perception, MPC planning, and low-level PD control modules communicate via ROS 2 interfaces.

The MPC planner operates at 25\,Hz ($T_s=40\,\text{ms}$), formulated in CasADi \cite{andersson2019casadi} and solved by IPOPT \cite{nocedal2009adaptive} with the MA57 solver. It outputs joint references to 1\,kHz PD controllers on the hardware. Code generation feature was employed to enhance execution performance. 

\subsection{Experimental Results and Discussions}
\subsubsection{Real-World Catching Demonstration} 
\label{subsubsec:real_world_demo}

To evaluate the performance of the overall framework on the studied system, we conducted a series of 35 consecutive real-world throws using the proposed AT-MPC framework. Out of these attempts, the system successfully intercepted 13 throws ($37.1\%$ success rate), and 3 attempts ($8.6\%$) were automatically aborted due to estimated actuator power limit violations. The remaining failures were primarily attributed to external factors, such as camera perception instability or throws landing outside the predefined safe interception zone.

To demonstrate the efficacy and real-time feasibility of the proposed MPC framework, a representative successful catching trial using the AT mode is presented. Fig.~\ref{fig:Fig_3D_Catch} visualizes the trajectory generation process during this trial. Initially, predictions from the Kalman filter may fall outside the predefined safe workspace due to the limited number of early measurements. Once the predicted trajectory intersects this designated safe zone, the Point Selection Algorithm determines a feasible interception target pose $p_{tg}$. This target is then supplied to the planner, which computes kinematically feasible joint reference trajectories $\{q_d(k), \dot{q}_d(k)\}_{k=0}^{N-1}$ for both manipulators.

The resulting motion execution is shown by the actual catcher trajectory (blue line). Fig.~\ref{fig:Fig_3D_Catch Close} offers a close-up view, highlighting the close spatial correspondence between the MPC-generated reference and the trajectory executed by the low-level PD controllers. Qualitatively, this minimal deviation between the planned path and the executed path in 3D space indicates effective tracking performance by the low-level PD control, validating the overall hierarchical control structure in Fig.~\ref{fig:system diagram}. Minor observed spatial discrepancies can be primarily attributed to factors including the inherent limitations of PD control in perfectly tracking high-speed, dynamic references, potential model inaccuracies, and communication latencies within the ROS 2 framework.

Crucially, manipulator coordination and closed-chain constraint satisfaction were verified. To quantify this, we measured the Euclidean distance between the two end-effector grasping points throughout the experiments. Theoretically, this distance should remain constant at its nominal value, $d_{\text{nom}}$, as our NMPC formulation explicitly enforces (\ref{eq:closed_chain_constraint}) required by the rigid grasp. The experimental deviation from $d_{\text{nom}}$ remained minimal throughout the maneuver, averaging only 2.8mm with a worst-case of 6.1mm. Minor fluctuations are likely attributed to contact compliance and low-level tracking synchronization effects, not a constraint violation within the planner itself.

\begin{figure}[t]
    \centering
    \includegraphics[width=0.9\columnwidth]{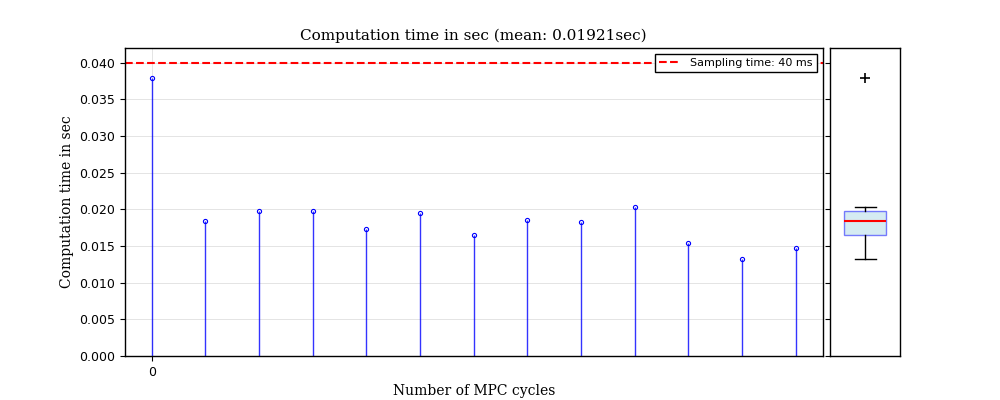}
    \caption{Real-time performance: End-to-end computation times per AT-MPC cycle during a real-world experiment (cf. Fig.~\ref{fig:Fig_3D_Catch}). Includes full node processing from state input to reference output for the MPC node. Average ($19.21\,\text{ms}$) and worst-case times remained well below $T_s=40\,\text{ms}$.}
    \label{fig:Fig_timing_realcatch}
    \vspace{-2mm}
\end{figure}

Furthermore, real-time computational performance was validated and shown in Fig.~\ref{fig:Fig_timing_realcatch}. The average MPC cycle computation time was $19.21\,\text{ms}$, significantly below the sampling time $T_s$. The worst-case computation time (excluding potential initial overhead) also remained comfortably within this budget, demonstrating the planner's real-time feasibility and capability for consistent plan updates during the dynamic task.

\subsubsection{Comparative Analysis of MPC Modes (AT vs. PT)}
\label{subsec:comparison}

During extensive real-world catching trials, a notable limitation was observed when employing the Primitive-Terminal (PT) mode in (\ref{eq:mpc_cost_primitive}). In scenarios demanding agile and significant configuration adjustments, the PT planner frequently generated trajectories that triggered exceptions related to estimated power limit violations within the Franka Control Interface (FCI)\footnote{https://github.com/frankarobotics/libfranka/issues/143}, leading to task abortion. This phenomenon was significantly less prevalent in trials using the AT mode. We hypothesize that the PT mode's fixed, aggressive penalization of terminal state errors can lead the MPC to compute overly assertive joint acceleration commands $\ddot{q}(k)$, particularly early in the trajectory when errors are large. As our MPC formulation is based on a kinematic model in (\ref{eq:mpc_ss_model}) without explicit consideration of dynamic constraints (e.g., torque or power limits), these commands, when executed by the low-level controllers, may exceed hardware capabilities, especially in certain dynamically challenging configurations.

To quantitatively investigate this phenomenon and compare the intrinsic performance characteristics of the AT and PT planners in a controlled setting, we conducted a comparative analysis in simulation. This approach ensures repeatability by allowing identical throwing conditions (represented as start and target pose pairs for the planner) for both modes, while also isolating the planner's performance from hardware noise, communication latencies, and low-level control artifacts. For this purpose, $N=500$ trials were executed, each with randomly generated start and target poses within the workspace, ensuring that for each trial, both AT and PT modes received the exact same planning task.

\begin{figure}[t]
    \centering
    \includegraphics[width=0.9\columnwidth]{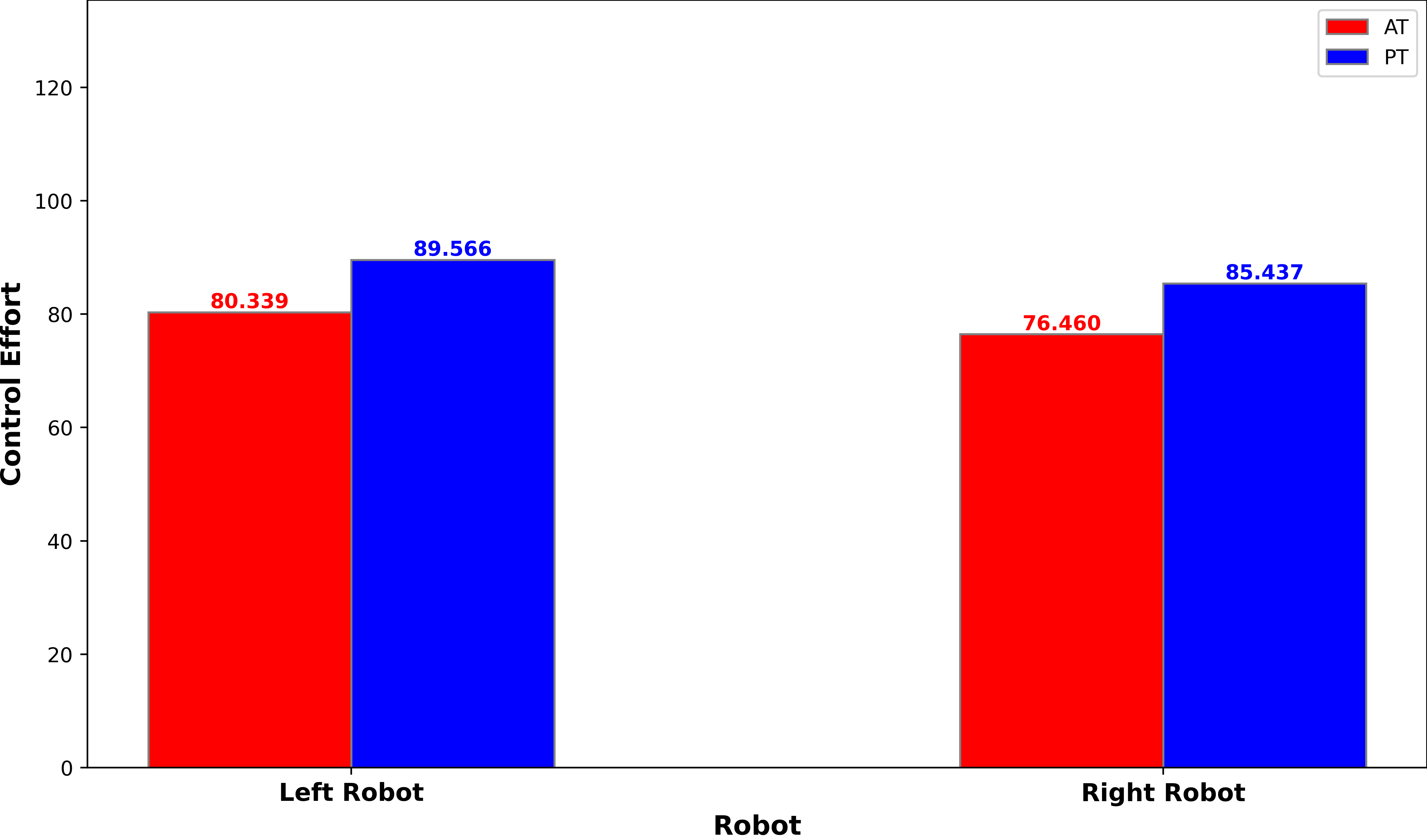}
    \caption{Average control effort ($E = \sum \|\ddot{\bm{q}}(k)\|_2^2$) for AT vs. PT modes across 500 trials. The AT mode demonstrates significantly reduced effort for both manipulators.}
    \label{fig:Fig_ControlEffort}
    \vspace{-2mm}
\end{figure}

Control effort, defined as the sum of squared joint accelerations over the entire duration of the planned trajectory (i.e., from start to target interception): $E = \sum_{k=0}^{N_{traj}} \|\ddot{\bm{q}}(k)\|_2^2$, was evaluated as a primary metric, serving as a proxy for energy consumption and the likelihood of triggering power limits. Fig.~\ref{fig:Fig_ControlEffort} presents the average control effort computed over 500 simulated trials for both the AT and PT modes, initiated from randomized start/target poses within the workspace.

The results indicate that the AT mode required significantly lower average control effort compared to the PT mode for both manipulators. This quantitative finding supports our hypothesis and real-world observations regarding power limit exceptions. The adaptive weighting mechanism inherent to the AT mode in (\ref{eq:mpc_cost_adaptive}) tends to produce smoother acceleration profiles by moderating commands based on the error magnitude, thereby reducing peak actuator demands compared to the fixed, aggressive terminal penalization of the PT approach in (\ref{eq:mpc_cost_primitive}).

\begin{figure}[t]
    \centering
    \includegraphics[width=0.75\columnwidth]{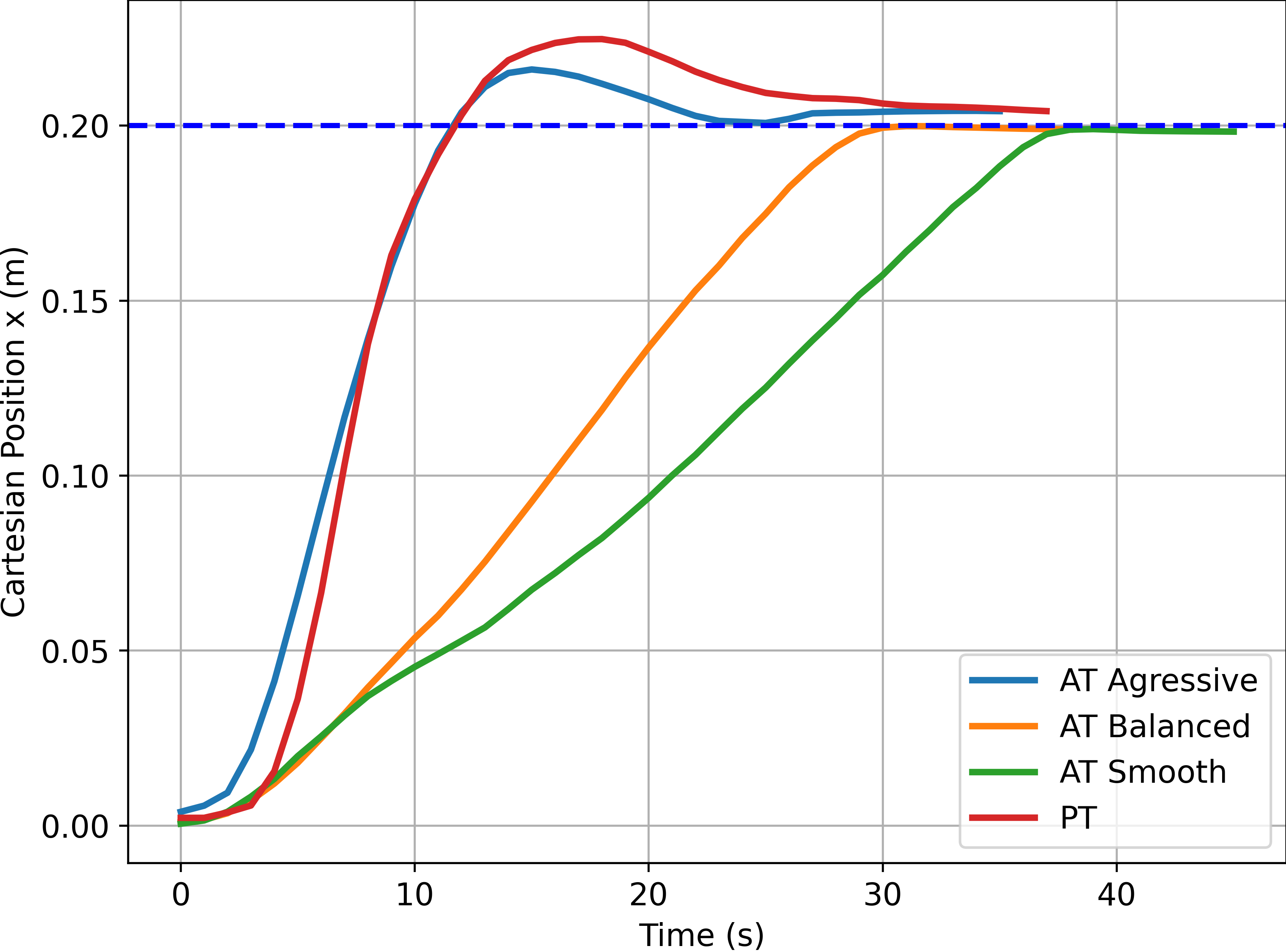}
    \caption{Comparison of Cartesian X-axis trajectory profiles for PT and differently tuned AT modes (Aggressive, Balanced, Smooth) in a representative simulation. Parameters are detailed in Table~\ref{tab:mpc_mode_params_revised_textsub}.}
    \label{fig:Fig_CMP_ATPT}
\end{figure}

Fig.~\ref{fig:Fig_CMP_ATPT} qualitatively compare of the Cartesian X-trajectory profiles from a representative trial for the PT mode alongside AT modes tuned for aggressive, balanced, and smooth responses. The specific parameter configurations differentiating these modes are detailed in Table~\ref{tab:mpc_mode_params_revised_textsub}. As hypothesized, the PT mode and the aggressively tuned AT mode (approximating PT's high terminal weights) exhibit rapid convergence but suffer from considerable overshoot. In contrast, a balanced or smoother AT weighting yields a more gradual approach with significantly reduced overshoot. Specifically, the AT Smooth configuration utilizes a near-zero adaptive terminal weight range, effectively minimizing the influence of the terminal cost terms in (\ref{eq:mpc_cost_adaptive}). This approximates a scenario prioritizing trajectory smoothness, primarily governed by stage costs, over strict terminal accuracy. This visually demonstrates the flexibility afforded by the AT formulation in tuning the trade-off between convergence speed and trajectory smoothness via its adaptive parameters, particularly the range $[W_{\min}, W_{\max}]$ governing the stage cost adaptation.

\begin{figure}[t]
    \centering
    \includegraphics[width=0.8\columnwidth]{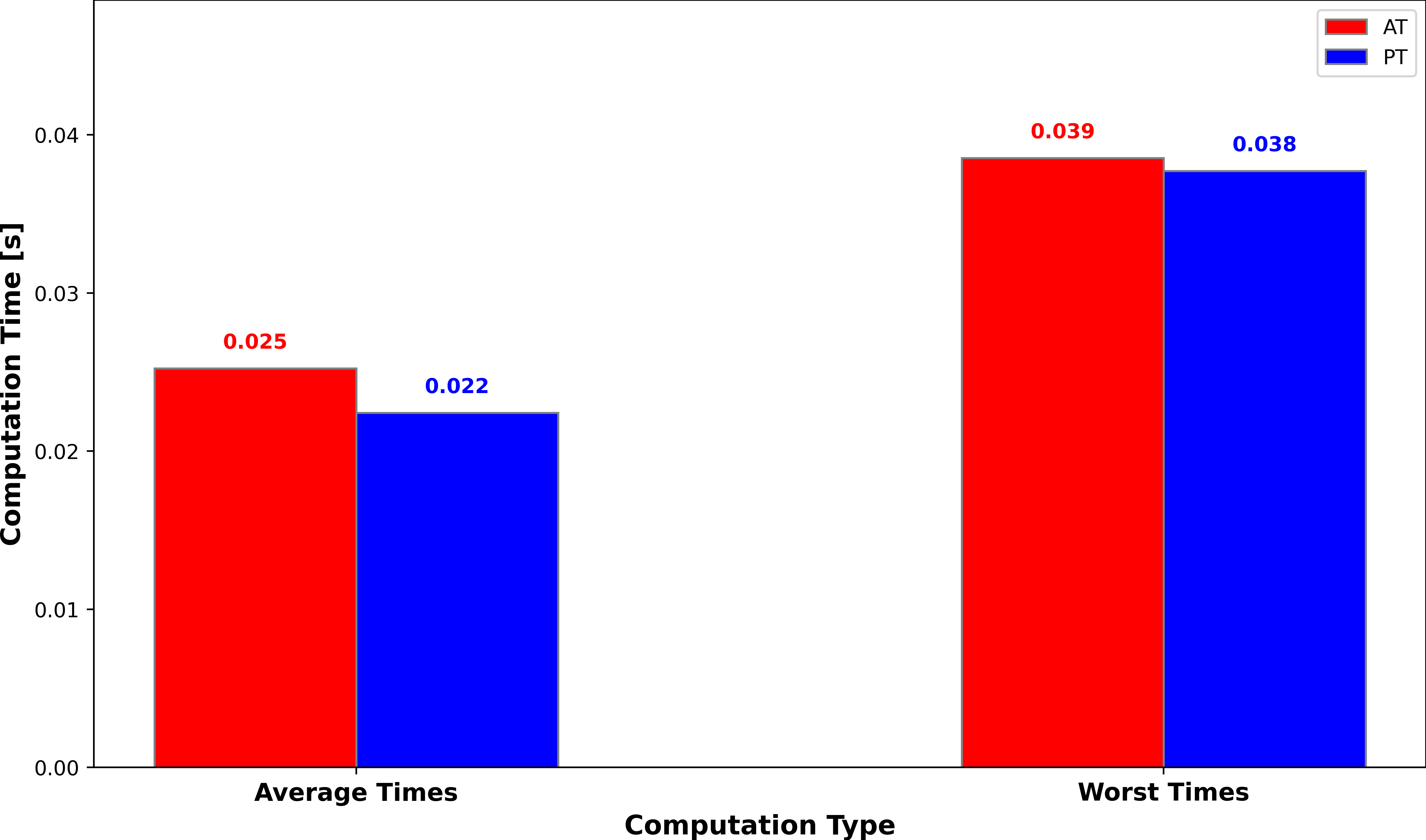}
    \caption{Computation time statistics over 500 trials. The bars display average (left) and worst-case (right) MPC cycle times for AT and PT modes. All times remain well below the $40\,\text{ms}$ sampling limit.}
    \label{fig:Fig_timingStatistics_ATPT}
    \vspace{-1mm}
\end{figure}

\begin{figure}[t]
    \centering
    \includegraphics[width=0.9\columnwidth]{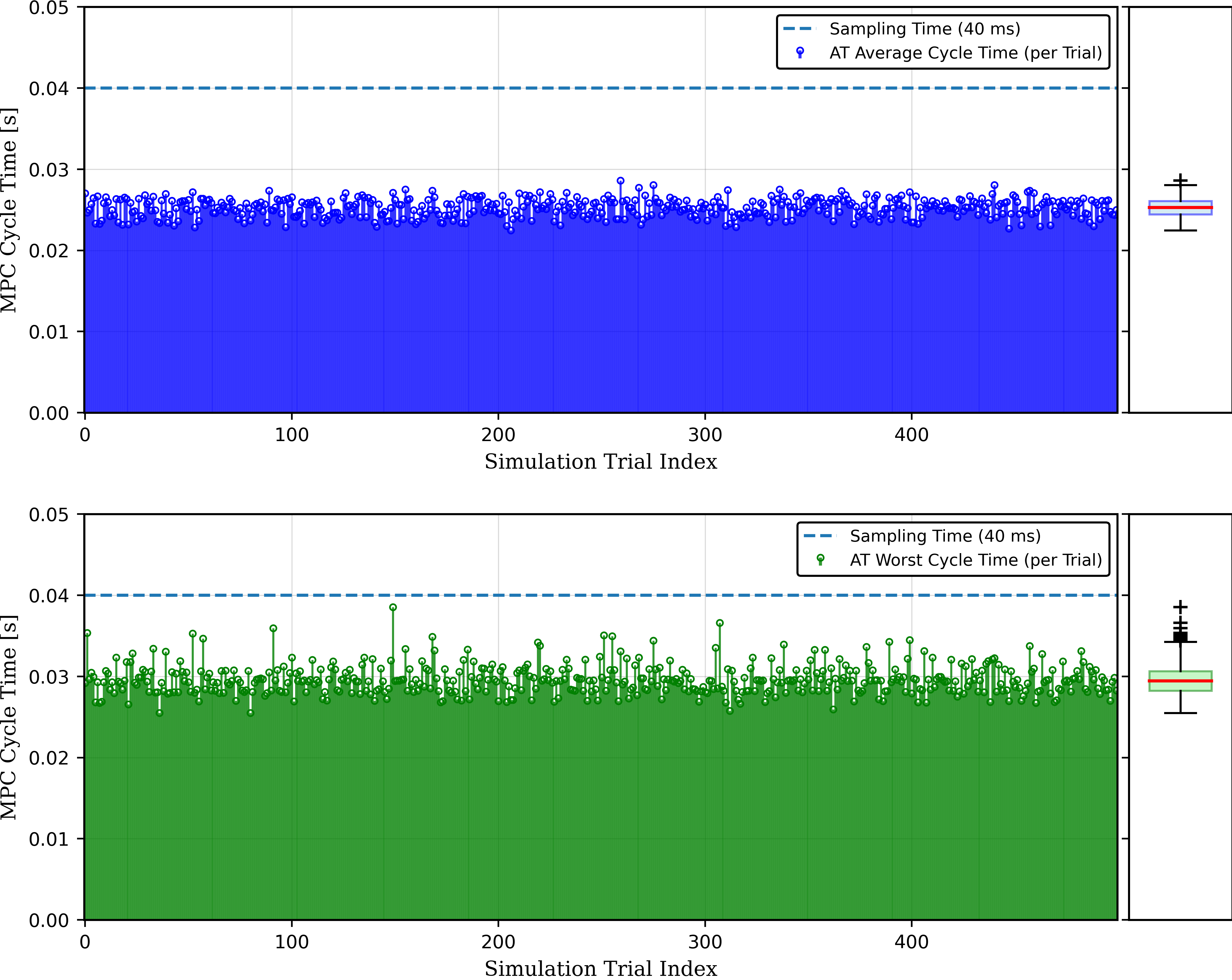}
    \caption{Computation time distributions for AT (Balanced) mode across 500 trials. \textbf{Top/Bottom:} Per-trial average (blue) and worst-case (green) cycle times with aggregate box plots. All instances remain within the $40\,\text{ms}$ real-time limit (dashed line).}
    \label{fig:Fig_timingStatistics_ATBox}
    \vspace{-1mm}
\end{figure}

\begin{table}[h!]
    \vspace{0.3mm}
    \centering
    \caption{Parameter configurations differentiating the MPC modes compared in Fig.~\ref{fig:Fig_CMP_ATPT}. PT uses fixed terminal weights. AT variants differ in their adaptive terminal weight ranges. \textbf{All AT variants share the same adaptive stage weight ranges:} Position $(Q_{pos,\text{max}}^{adp}, Q_{pos,\text{min}}^{adp}) = (2.0, 0.1)$ and Orientation $(Q_{ori,\text{max}}^{adp}, Q_{ori,\text{min}}^{adp}) = (1.0, 0.1)$.}
    \label{tab:mpc_mode_params_revised_textsub} 
    \renewcommand{\cellalign}{cl} 
    \setcellgapes{3pt} 
    \makegapedcells
    \begin{tabular}{l l}
    \toprule
    \textbf{MPC Mode} & \textbf{Parameters} \\
    \midrule
    PT & $P_e = 500$, $O_e = 10$ \\
    \addlinespace 
    
    AT Aggressive & \makecell[l]{
                        $(P_{e,\text{max}}^{adp}, P_{e,\text{min}}^{adp}) = (500, 100)$ \\
                        $(O_{e,\text{max}}^{adp}, O_{e,\text{min}}^{adp}) = (10, 2)$
                    } \\
    \addlinespace
    
    AT Balanced & \makecell[l]{
                        $(P_{e,\text{max}}^{adp}, P_{e,\text{min}}^{adp}) = (50, 20)$ \\
                        $(O_{e,\text{max}}^{adp}, O_{e,\text{min}}^{adp}) = (10, 2)$
                    } \\
    \addlinespace
    
    AT Smooth & \makecell[l]{
                        $(P_{e,\text{max}}^{adp}, P_{e,\text{min}}^{adp}) = (0.001, 0)$ \\ 
                        $(O_{e,\text{max}}^{adp}, O_{e,\text{min}}^{adp}) = (0.001, 0)$  
                    } \\
    \bottomrule
    \end{tabular}
\end{table}
Finally, the computational efficiency of both modes was compared over the 500 simulation trials (Fig.~\ref{fig:Fig_timingStatistics_ATPT}). The AT formulation's adaptive logic introduced minimal computational overhead; its average time was only marginally higher than the PT's, and the worst-case times were comparable ($39\,\text{ms}$ vs $38\,\text{ms}$) . As illustrated in Fig.~\ref{fig:Fig_timingStatistics_ATBox}, the AT mode also demonstrated high consistency, with both its average and worst-case cycle times remaining well below the $T_s = 40\,\text{ms}$ across all 500 scenarios.

These simulation results confirm that both modes meet the real-time constraints. Crucially, they demonstrate that the additional complexity of the AT mode introduces minimal computational overhead, reinforcing its practical applicability as demonstrated in the physical experiments in Section~\ref{subsubsec:real_world_demo}. 

\section{CONCLUSIONS}
\label{sec:conclusions_future_work_condensed}

This paper presented and experimentally validated an MPC-based motion planner, demonstrating the effectiveness of an Adaptive-Terminal (AT) formulation for cooperative ball catching by two manipulators under closed-chain constraints. Compared to a baseline Primitive-Terminal (PT) approach, the AT strategy significantly reduced control effort and mitigated actuator power limit violations, yielding smoother trajectories with minimal computational overhead, confirmed by average cycle times well below the $40\,\text{ms}$ sampling period. Future work will focus on enhancing robustness and success rates through a learning-augmented hierarchical framework, training policies to predict optimal interception poses directly from observations, aiming for improved overall system performance in dynamic cooperative tasks.

\bibliography{refs}             
                                                   
\end{document}